\documentclass[lettersize,journal]{IEEEtran}
\usepackage{amsmath,amsfonts,amssymb}
\usepackage{algorithmic}
\usepackage{algorithm}
\usepackage{array}
\usepackage{booktabs}
\usepackage[caption=false,font=normalsize,labelfont=sf,textfont=sf]{subfig}
\usepackage{textcomp}
\usepackage{stfloats}
\usepackage{url}
\usepackage{verbatim}
\usepackage{graphicx}
\usepackage{capt-of}
\graphicspath{{figures/}}
\usepackage{cite}
\begin{document}

\title{DPSF-Net: A Dual-Prior Spatial-Frequency Network for Real-World Remote Sensing Image Dehazing}

\author{
Mei Lu,
Shangliang Shao,
Shanliang Yao,

% \thanks{This work was supported in part by the Scientific Research Foundation for Advanced Talents (No. jit-b-202045), and in part by the Jiangsu Provincial Industry-University-Research Cooperation Project (No. BY20230040).}

\thanks{$^{\text{1}}$ Mei Lu and Shangliang Shao are with the School of Software Engineering, Jinling Institute of Technology, Nanjing 211100, China. (e-mail: mlu@jit.edu.cn; shaoshangliang182@163.com). }

\thanks{$^{\text{2}}$ Shanliang Yao is with the School of Information Engineering, Yancheng Institute of Technology, Yancheng 224051, China. (email: shanliang.yao@ycit.edu.cn).}

}

\maketitle

\begin{abstract}
Real-world remote sensing image dehazing (RSID) remains challenging because atmospheric scattering, spatially non-uniform haze and colour distortion jointly degrade structural and spectral information. Most deep learning methods rely on RGB inputs and spatial-domain feature extraction, which limits their ability to separate global background haze from local surface details. Here, we propose DPSF-Net, a dual-prior spatial-frequency network built on MCAF-Net for real-world RSID. The network uses hazy RGB images and dark channel prior (DCP) maps as joint inputs, allowing physical degradation cues to guide end-to-end feature learning. A spatial-frequency residual interaction block introduces a FourierUnit branch into multi-directional spatial interaction to model large-scale haze components. A prior-guided feature attention module adaptively fuses prior and attention features to reduce colour shift and structural distortion. A selective kernel complementary fusion module screens multi-scale skip features through bidirectional residual complementary gating and selective kernel fusion. Extensive experiments demonstrate that DPSF-Net achieves state-of-the-art performance on the real-world RRSHID remote sensing image dehazing benchmark and remains competitive across multiple synthetic datasets. Moreover, the proposed method strikes a favourable balance among restoration quality, parameter count and computational complexity, supporting the effectiveness of dual-prior spatial-frequency modelling.
\end{abstract}


\section{Introduction}

Remote sensing images support urban planning, agricultural monitoring, environmental assessment, disaster response, target detection and scene classification \cite{ref1}, \cite{ref2}. During imaging, water vapour, aerosols and suspended particles scatter and absorb radiation reflected from ground objects \cite{ref3}, \cite{ref4}. The resulting haze reduces contrast, blurs edges, attenuates texture and shifts colour. These degradations lower visual quality and can also compromise downstream tasks such as semantic segmentation, change detection and object recognition. Robust dehazing for real-world remote sensing scenes is therefore both scientifically and practically important.

% Queue the single-column overview after the opening text so IEEEtran places it
% at the top of the first page's right-hand column rather than above the abstract.
\begin{figure}[!t]
\centering
\includegraphics[width=\linewidth]{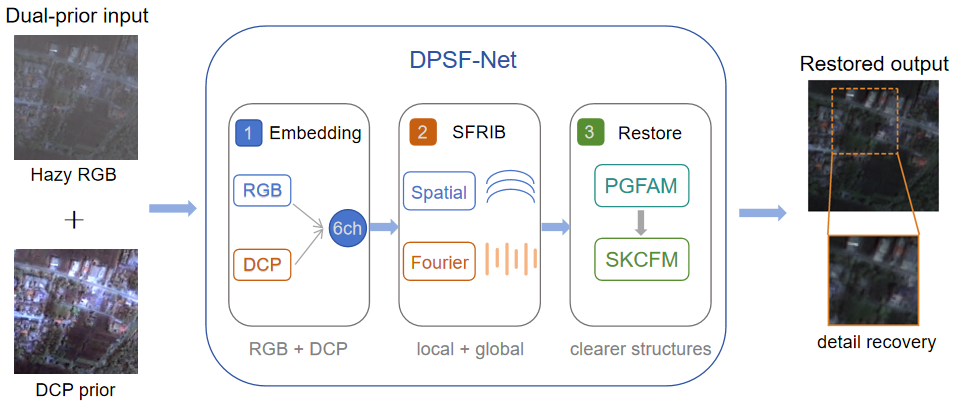}
\caption{Overview of DPSF-Net. The network combines hazy RGB imagery and DCP guidance to jointly model local structures and global haze distributions, producing clearer remote sensing images with improved detail recovery.}
\label{fig:overview}
\end{figure}

Early RSID methods estimated transmission maps or atmospheric light from physical imaging models and hand-crafted priors, which gave them useful interpretability \cite{ref5,ref6,ref7,ref8,ref9,ref10}. Their fixed assumptions, however, often break down over bright land cover, cloud-haze mixtures and heterogeneous surfaces. Deep learning has improved dehazing performance \cite{ref11,ref12,ref13,ref14,ref15,ref16,ref17,ref18}, but many networks remain trained mainly on synthetic data and use only RGB inputs. This limits their cross-domain generalisation to real scenes with non-uniform haze, complex scattering and sensor-dependent colour responses.

MCAF-Net provides an important benchmark for real-world RSID by introducing the RRSHID dataset and three modules for multi-dimensional spatial feature extraction, colour calibration and multi-scale feature fusion \cite{ref19}. Despite its strong performance, three limitations remain. First, MFIBA mainly models spatial and channel interactions, and its pointwise branch does not explicitly represent global low-frequency haze. Second, CSAM generates attention from internal features without direct physical prior guidance, which can weaken calibration under complex haze. Third, MFAFM fuses skip features adaptively, but shallow skip features can still carry residual haze into the decoder.

Real haze degradation in remote sensing imagery has coupled spatial and frequency characteristics \cite{ref20}. Spatially, haze reduces local contrast and blurs edges. In the frequency domain, atmospheric scattering strengthens smooth low-frequency components and suppresses high-frequency detail. Spatial convolution alone is therefore insufficient for recovering both local structures and global haze distributions. Although the dark channel prior is not always valid over bright remote sensing regions, it can still provide coarse cues about haze concentration and transmission variation \cite{ref7}. Used as an auxiliary input rather than a hard constraint, DCP can guide a deep network while preserving data-driven representation learning.

Motivated by these observations, we propose DPSF-Net, a dual-prior spatial-frequency network for real-world RSID. The model uses a 6-channel RGB-DCP input and combines data-driven visual features with physical prior information at the embedding stage. MFIBA is replaced by a spatial-frequency residual interaction block (SFRIB), which adds a FourierUnit branch to the HW, CH and CW spatial interaction branches to capture global haze degradation. CSAM is extended to a prior-guided feature attention module (PGFAM), which fuses prior and attention features for colour and structure calibration. MFAFM is replaced by a selective kernel complementary fusion module (SKCFM), which uses bidirectional residual complementary gating (BRCM) and selective kernel fusion (SKFusion) to suppress redundant haze in skip connections.

The main contributions are summarised as follows:

\begin{enumerate}

\item We propose DPSF-Net, an end-to-end dual-prior spatial-frequency network for real-world RSID. The framework integrates hazy RGB imagery with DCP guidance at the embedding stage, allowing physical degradation cues to complement data-driven feature learning.

\item We redesign the principal representation and fusion components of MCAF-Net. SFRIB combines multi-directional spatial interaction with Fourier-domain modelling, PGFAM adaptively merges prior-guided and attention features, and SKCFM uses complementary gating and selective kernel fusion to limit the propagation of residual haze through skip connections.

\item Extensive experiments on RRSHID and four public benchmarks show that DPSF-Net improves restoration quality while retaining a compact model size and moderate computational cost. Component-wise and loss-function ablations further clarify the contribution of each design choice.

\end{enumerate}

The remainder of this paper is organised as follows. Section II reviews physics-based RSID methods, deep dehazing networks and spatial-frequency modelling strategies. Section III presents the architecture of DPSF-Net and its optimisation objectives. Section IV describes the experimental protocol, reports quantitative and qualitative comparisons, and discusses the scope and limitations of the method. Section V concludes the study.

\section{Related work}

\subsection{Physics-based RSID methods}

Physics-based RSID methods generally estimate transmission maps and atmospheric light from the atmospheric scattering model \cite{ref3}, \cite{ref4}. Representative single-image dehazing approaches include local statistical transmission estimation \cite{ref5}, fast visibility restoration \cite{ref6}, the dark channel prior \cite{ref7}, colour attenuation prior \cite{ref8}, non-local prior \cite{ref9} and haze-line prior \cite{ref10}. These methods improve visibility under certain assumptions, but their fixed priors are vulnerable to bright land cover, cloud-haze overlap, complex surfaces and colour shifts in remote sensing imagery.

The dark channel prior remains informative because it offers a coarse description of haze concentration and transmission variation \cite{ref7}. Its reliability, however, decreases over bright roofs, bare soil, cloud boundaries and other high-reflectance regions that frequently occur in remote sensing scenes. Directly enforcing the prior may therefore amplify colour bias or remove useful surface responses. In this study, DCP is used as an auxiliary physical modality rather than a deterministic restoration rule. The network learns when the prior is informative and how strongly it should influence the restored representation.

Prior-based methods remain valuable even when their assumptions are imperfect. They expose physically meaningful variables and help distinguish atmospheric degradation from surface appearance. This interpretability is especially useful in remote sensing, where scenes contain heterogeneous materials and broad illumination variation. However, the same diversity makes a fixed prior difficult to apply uniformly. A practical design should retain the coarse atmospheric cue while allowing data-driven features to correct local violations of the prior. This consideration motivates the dual-prior embedding used in DPSF-Net.

\subsection{Deep learning-based RSID methods}

Deep learning methods learn the nonlinear mapping between hazy and clear images from data, reducing the need for manual prior estimation. Early networks for natural images included DehazeNet \cite{ref11}, MSCNN \cite{ref12} and AOD-Net \cite{ref13}. Subsequent architectures such as FFA-Net \cite{ref14}, GridDehazeNet \cite{ref15}, DehazeFormer \cite{ref16}, TransWeather \cite{ref17} and 4KDehazing \cite{ref18} improved feature fusion, multi-scale context modelling, long-range dependency capture and high-resolution restoration.

Natural-image dehazing and RSID share a common restoration objective, but their operating conditions differ. Remote sensing images cover larger spatial extents, and the same scene may contain roads, buildings, vegetation, water and bare soil. Haze is often spatially uneven and can overlap with thin clouds or sensor-dependent colour shifts. Consequently, a model that performs well on natural photographs may still struggle to preserve small targets and surface boundaries in aerial scenes. Effective RSID networks need both broad contextual perception and careful reuse of shallow structural features.

Remote sensing dehazing has since developed specialised architectures. Trinity-Net combines gradient guidance with Swin Transformer \cite{ref21}, RSDehazeNet introduces channel refinement for multispectral images \cite{ref22}, and HyperDehazing investigates hyperspectral information \cite{ref23}. PSMB-Net \cite{ref24}, DR3DF-Net \cite{ref25}, PCSformer \cite{ref26}, RSDehamba \cite{ref27}, HDMba \cite{ref28}, SCANet \cite{ref29} and PhDNet \cite{ref30} address non-uniform haze, cross-scale fusion, long-range modelling or physical constraints.

These studies show that RSID benefits from representations tailored to heterogeneous land-cover responses and large image extents. Nevertheless, several architectures still rely primarily on RGB inputs and spatial-domain operators. This design can limit sensitivity to global haze distributions and may allow degraded shallow features to propagate through decoder skip connections. MCAF-Net recently established a real-world RSID benchmark and baseline modules for spatial feature extraction, colour calibration and adaptive fusion \cite{ref19}. DPSF-Net extends this line of work by integrating explicit RGB-DCP guidance, Fourier-domain context modelling and selective skip-feature screening within a compact encoder-decoder framework.

The design of DPSF-Net follows a conservative extension strategy. Instead of replacing the lightweight baseline with a substantially larger architecture, we revise the stages most closely related to real-world degradation. The input stage receives an auxiliary physical cue, the representation stage gains a frequency-domain branch, and the decoder screens skip features before reuse. This approach makes the source of each performance change easier to evaluate through ablation studies.

\subsection{Spatial-frequency modelling and dynamic fusion}

Haze degradation appears differently across spatial and frequency domains. Spatially, it blurs edges and weakens texture. In frequency space, it strengthens low-frequency haze components and attenuates high-frequency detail. Pure spatial convolution may therefore miss large-scale non-uniform haze, whereas isolated frequency processing can overlook local structural relationships.

Recent Fourier, wavelet and spatial-frequency fusion methods have been used to capture global degradation cues \cite{ref31,ref32,ref33,ref34}. Fast Fourier convolution provides an efficient route for frequency-domain context modelling \cite{ref35}, and selective kernel mechanisms dynamically adjust branch or scale contributions \cite{ref36}. SFRDP-Net further demonstrates the value of spatial-frequency residual guidance for remote sensing image dehazing \cite{ref20}.

Frequency-domain processing and dynamic fusion address different aspects of the restoration problem. Fourier operators provide image-wide context with limited additional depth, which is useful for spatially extended haze. Selective fusion adjusts the contribution of multi-scale features according to their channel responses and scene content \cite{ref36}. Because skip connections may transmit both useful details and residual degradation, these mechanisms are complementary. DPSF-Net combines them with prior-guided attention to improve global haze perception, local detail recovery and redundant-response suppression.

An important distinction is that spatial and frequency representations are not interchangeable. Spatial interaction is well suited to local boundaries, directional textures and small objects. Fourier-domain processing provides a complementary view of smooth variations that extend across a wider region. For RSID, these variations often correspond to non-uniform background haze. A joint design can therefore preserve local geometry while improving awareness of large-scale degradation. Dynamic fusion then determines which multi-scale responses should be retained during reconstruction.

\section{Proposed method}

The overall architecture of DPSF-Net is shown in Fig. 2. Following MCAF-Net \cite{ref19}, the network adopts a U-shaped encoder-decoder framework \cite{ref37}, but redesigns the input, feature extraction, attention and skip-fusion stages for real-world haze. Hazy RGB images and their DCP maps are concatenated into a 6-channel input and embedded through separate appearance and prior branches. The encoder uses spatial-frequency residual interaction blocks (SFRIBs) to combine directional spatial modelling with Fourier-domain context. Prior-guided feature attention modules (PGFAMs) calibrate intermediate representations before decoder fusion. Selective kernel complementary fusion modules (SKCFMs) then refine multi-scale skip features using bidirectional residual complementary gating and selective kernel fusion. A residual reconstruction layer finally predicts the restored RGB image. This design preserves the lightweight backbone while improving sensitivity to non-uniform haze, colour deviation and weak surface details.

The revised architecture is guided by three principles. First, physical information should assist representation learning without acting as a hard restoration constraint. Second, global haze perception should be added without discarding the local modelling capacity of the baseline. Third, decoder fusion should retain useful structural detail while limiting the return of shallow degradation responses. These principles correspond to dual-prior embedding, SFRIB and the combined PGFAM-SKCFM reconstruction pathway.

\begin{figure*}[!t]
\centering
\includegraphics[width=\textwidth]{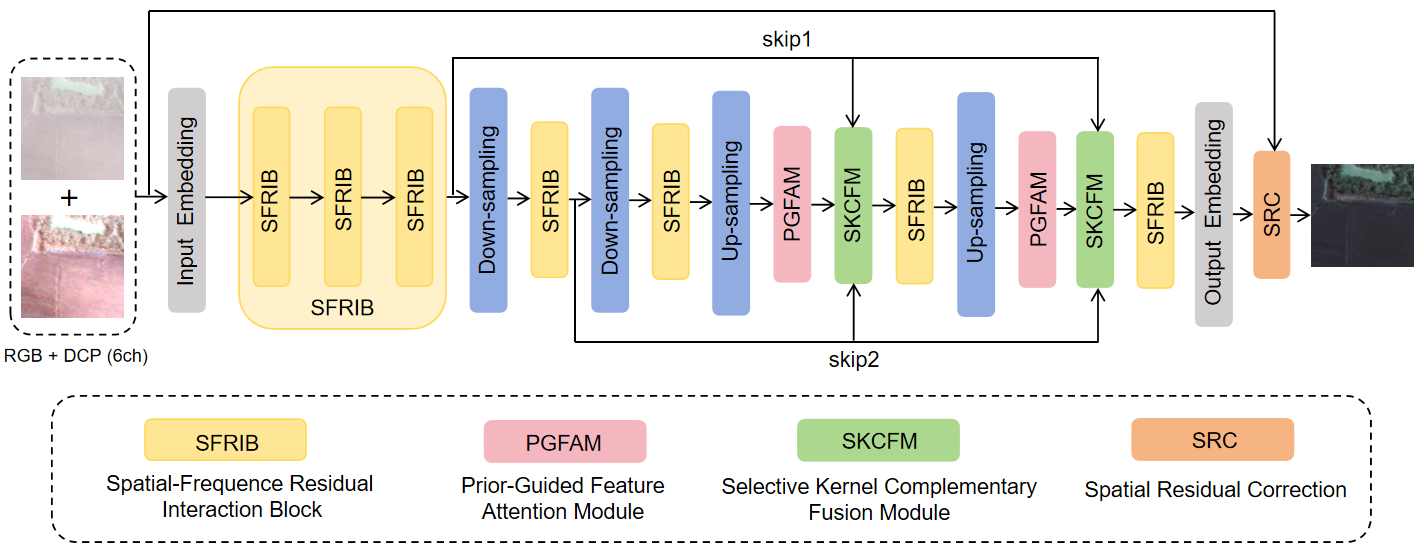}
\caption{Overall architecture of DPSF-Net. The encoder-decoder network uses dual-prior embedding, SFRIB-based spatial-frequency modelling, PGFAM-based prior-guided calibration and SKCFM-based selective skip-feature fusion.}
\label{fig:2}
\end{figure*}

\subsection{Dual prior and feature embedding}

To improve the network's perception of real haze degradation, DPSF-Net introduces a dual-prior input composed of RGB imagery and DCP. Given a hazy remote sensing image $I_{\mathrm{rgb}} \in \mathbb{R}^{3 \times H \times W}$, the corresponding DCP map $I_{\mathrm{dcp}} \in \mathbb{R}^{3 \times H \times W}$ is first computed. The 6-channel input is then obtained by channel-wise concatenation:

\begin{equation}
X_{\mathrm{in}} = \operatorname{Concat}(I_{\mathrm{rgb}}, I_{\mathrm{dcp}}).
\end{equation}

During PatchEmbed, the 6-channel input is not mapped by a single convolution. Instead, separate RGB and DCP branches extract visual appearance features and physical prior features, which are concatenated and fused by a $1 \times 1$ convolution. This arrangement prevents the prior map from being treated as an ordinary colour channel and allows the network to learn a balanced early representation. Texture, edge and haze-distribution cues are therefore introduced before hierarchical encoding. DCP remains auxiliary guidance throughout the network, and the training losses are computed only on the restored RGB image.

The separation of the two branches is deliberate. RGB channels describe observed appearance, whereas DCP channels summarise a prior response derived from local intensity statistics. Their distributions and semantic roles are not identical. Independent projections allow each modality to develop an appropriate shallow representation before fusion. The subsequent $1 \times 1$ convolution learns channel interactions with limited overhead and supplies a unified feature tensor to the encoder.

\subsection{Spatial-frequency residual interaction block}

Real remote sensing haze contains local texture degradation and global low-frequency background components. MFIBA in MCAF-Net models HW, CH and CW interactions, but its fourth pointwise convolution branch does not explicitly capture global frequency-domain degradation \cite{ref19}. SFRIB replaces this branch with a FourierUnit \cite{ref35} while retaining the multi-directional spatial branches. The module structure is shown in Fig. 3.

Given an input feature $X \in \mathbb{R}^{B \times C \times H \times W}$, SFRIB first normalises $X$ and splits it into four channel groups:

\begin{equation}
X_1, X_2, X_3, X_4 = \operatorname{Split}(\operatorname{LN}(X)).
\end{equation}

$X_1$, $X_2$ and $X_3$ are fed into the HW, CH and CW spatial-interaction branches, respectively. These branches model correlations across the spatial plane, channel-height and channel-width directions. As in MFIBA, learnable parameters, depthwise convolution and dynamic weights enhance useful structures and suppress redundant haze responses.

For the fourth sub-feature $X_4$, SFRIB replaces the original pointwise convolution branch with FourierUnit-based frequency-domain modelling. The feature is mapped to the frequency domain by FFT, and the real and imaginary spectral components are concatenated along channels. A $1 \times 1$ convolution models frequency responses associated with global haze degradation. The enhanced spectrum is then restored to the spatial domain by iFFT and projected by a $1 \times 1$ convolution. This branch compensates for the limited receptive field of local convolution and improves large-scale background haze perception.

The outputs of the three spatial branches and the FourierUnit branch are concatenated, normalised and fused by a lightweight MLP. A residual connection then preserves the input representation:

\begin{equation}
Y = \operatorname{MLP}\left(\operatorname{LN}\left(\operatorname{Concat}(\widehat{X}_1,\widehat{X}_2, \widehat{X}_3, \widehat{X}_4)\right)\right) + X.
\end{equation}

Here, the fused representation combines three spatial-interaction outputs and one frequency-domain output. SFRIB keeps the same input and output dimensions ($B \times C \times H \times W$), while adding global haze perception to the local structure-modelling capacity of MFIBA. The directional branches retain sensitivity to anisotropic structures such as roads, field boundaries and building edges. In parallel, the FourierUnit provides a larger effective receptive field for smooth background degradation. Their residual integration limits excessive modification of well-preserved regions and supports stable optimisation.

The three spatial branches preserve the directional decomposition inherited from MFIBA. They model interactions over the spatial plane, channel-height plane and channel-width plane, respectively. Before these responses are fused, the network predicts dynamic weights from the current feature map. The weights allow the spatial contribution to vary with scene content. The fourth group is processed differently: FFT maps the feature into the spectral domain, a lightweight convolution modifies the concatenated real and imaginary responses, and iFFT restores the spatial layout. This branch is intended to complement the directional paths rather than replace them.

SFRIB is inserted throughout the five-stage backbone. Its residual form is important because dehazing should not alter every spatial region equally. In lightly degraded areas, the identity path can preserve existing content. In heavily degraded areas, the learned interaction branches provide additional correction. The block therefore supports a gradual refinement process across encoder and decoder stages while keeping the dimensional interface compatible with the baseline network.

\begin{figure}[t]
\centering
\includegraphics[width=\linewidth]{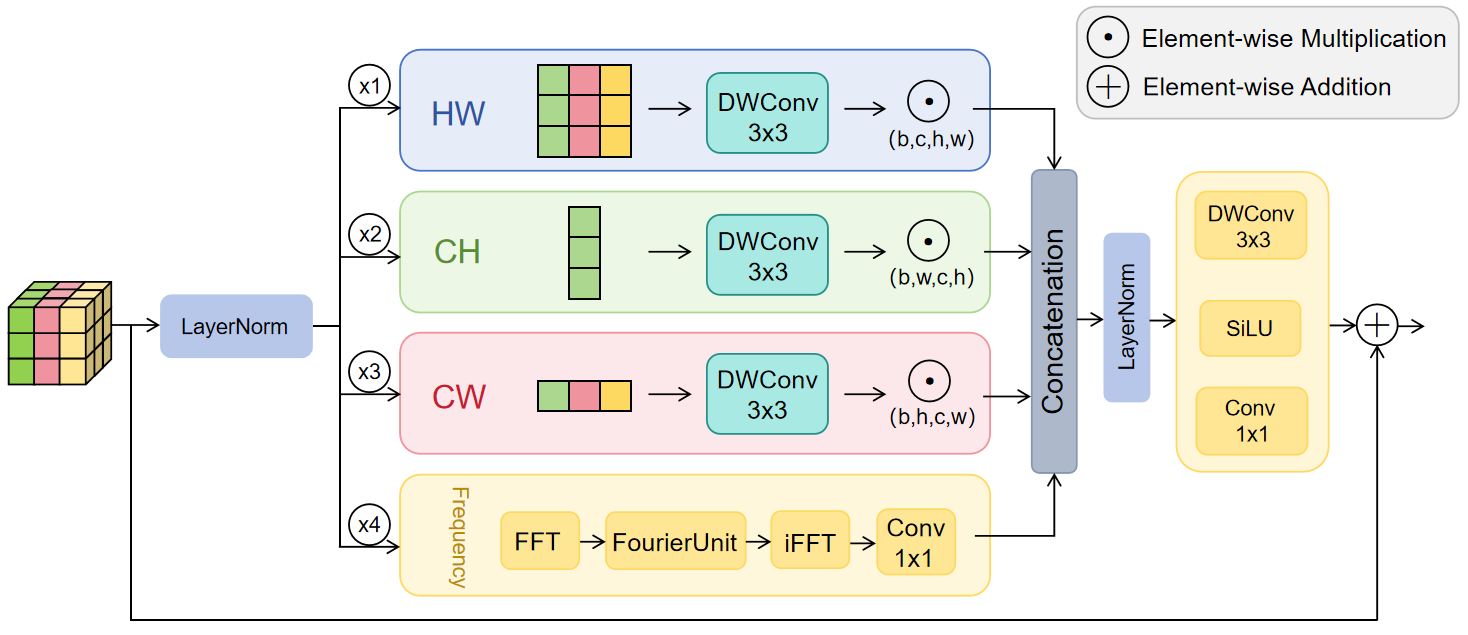}
\caption{Spatial-frequency residual interaction block (SFRIB). Three directional spatial branches model local structures, while the FourierUnit branch captures global haze distributions before residual fusion.}
\label{fig:3}
\end{figure}

\subsection{Prior-guided feature attention module}

Real remote sensing haze is often accompanied by colour shift and structural distortion. CSAM in MCAF-Net mitigates colour distortion by generating an intermediate pseudo-image and applying Q-K-V attention \cite{ref38}, but its attention relies mainly on internal network features. PGFAM extends CSAM with a prior-enhancement branch and an adaptive fusion mechanism, as shown in Fig. 4.

Given an input feature $X \in \mathbb{R}^{B \times C \times H \times W}$, PGFAM first uses PixelShuffle and a $1 \times 1$ convolution to generate a three-channel intermediate feature $F_{qk}$, which is converted into a pseudo-image $I_f$ by the colour-correction layer. $F_{qk}$ is then mapped to query $Q$ and key $K$ through LayerNorm and a $1 \times 1$ convolution, while $X$ is mapped to value $V$. After $L_2$ normalisation of $Q$ and $K$, the attention map is computed as follows:

\begin{equation}
A = \operatorname{Softmax}(QK^{\top} \cdot \alpha).
\end{equation}

where $\alpha$ is a learnable scaling parameter. The attention map $A$ is applied to $V$ and combined with the depthwise convolution branch of $V$ to obtain the attention-enhanced feature $F_{\mathrm{attn}}$.

PGFAM further introduces a prior-enhancement branch that extracts a prior representation $F_p$ from the current fusion feature $X$. The branch contains LayerNorm, a $3 \times 3$ convolution, ReLU and a $1 \times 1$ convolution. $F_p$ is not recalculated directly from the original DCP image. Instead, it is extracted from the pre-embedded RGB-DCP features, preserving the physical prior introduced at the input.

Finally, PGFAM concatenates $F_{\mathrm{attn}}$ and $F_p$ along the channel dimension and generates a fusion weight $W$ through a $1 \times 1$ convolution and a Sigmoid function:

\begin{equation}
F = F_{\mathrm{attn}} \odot W + F_p \odot (1-W).
\end{equation}

The fused feature is normalised and projected to produce the enhanced output. This adaptive balance between attention features and prior-guided features helps reduce colour deviation, structural ambiguity and uneven local recovery. Importantly, the fusion weight is estimated from the current representation rather than fixed globally. PGFAM can therefore attenuate unreliable prior responses over bright surfaces while retaining useful haze cues elsewhere. During training, the module also generates anchor features from pseudo-images and uses clear and hazy RGB images to compute an auxiliary contrastive loss.

PGFAM separates representation calibration from direct image restoration. The attention pathway estimates correlations from normalised query and key features, while the value pathway retains the feature content to be reweighted. A depthwise convolution supplements attention with a local response. In parallel, the prior-enhancement branch processes the embedded RGB-DCP representation using LayerNorm and convolutional projection. The final sigmoid weight balances these two pathways at each location and channel.

The pseudo-image branch also provides an auxiliary learning signal. During training, the projected anchor is encouraged to remain closer to the clear target than to the hazy input. This constraint is applied at two decoder scales. It does not alter the inference interface, but it encourages intermediate features to favour restoration-oriented representations. The module therefore combines adaptive inference-time fusion with an additional training-time regulariser.

\begin{figure}[t]
\centering
\includegraphics[width=\linewidth]{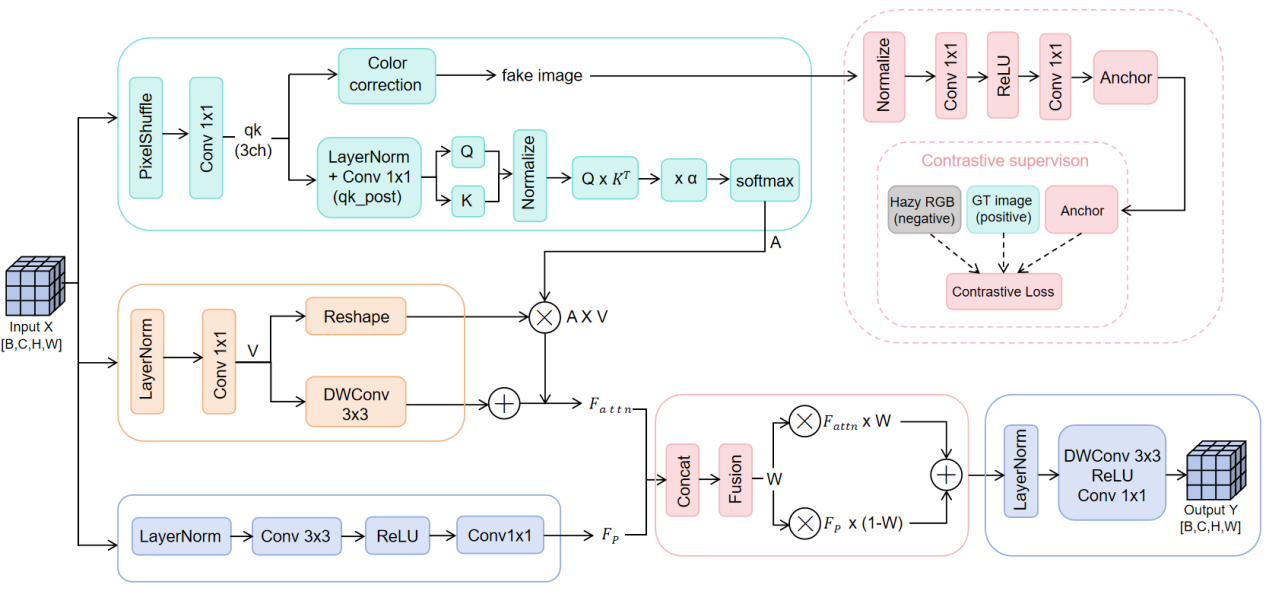}
\caption{Prior-guided feature attention module (PGFAM). Attention-enhanced features and embedded prior features are adaptively fused, and the pseudo-image branch provides two-scale contrastive supervision during training.}
\label{fig:4}
\end{figure}

\subsection{Selective kernel complementary fusion module}

Skip connections in U-shaped dehazing networks transmit shallow edge, texture and spatial details to the decoder \cite{ref37}. They may also pass residual haze, noise and redundant responses. MFAFM in MCAF-Net improves multi-scale fusion \cite{ref19}, but still has limited ability to model complementary relations between trunk and skip features. SKCFM addresses this issue with BRCM and SKFusion \cite{ref36}, as shown in Fig. 5.

Given the decoder trunk feature $X$ and the two encoder skip features $S_1$ and $S_2$, SKCFM first aligns $S_1$ and $S_2$ to the spatial resolution of $X$ and then aligns their channels using $1 \times 1$ convolutions. To reduce redundant haze transmission through skip connections, BRCM complements and enhances trunk and skip features in both directions:

\begin{equation}
\widetilde{S}_i = \operatorname{BRCM}(X, S_i), \quad i \in \{1,2\}.
\end{equation}

where $\widetilde{S}_1$ and $\widetilde{S}_2$ are complementary skip features guided by the trunk feature. BRCM uses gated residual information between trunk and skip features to enhance useful structures and suppress ineffective degradation responses.

SKCFM then constructs three multi-scale branches. The trunk feature $X$ is processed by a $3 \times 3$ convolution for current-scale context, while the enhanced $\widetilde{S}_1$ and $\widetilde{S}_2$ are processed by $5 \times 5$ and $7 \times 7$ convolutions to introduce larger receptive fields. After $3 \times 3$ refinement, SKFusion aggregates the three features and generates branch-channel weights through global average pooling, a lightweight MLP and Softmax. The final feature is obtained by weighted summation:

\begin{equation}
F = \sum_{i=1}^{3}\alpha_i F_i.
\end{equation}

where $F_i$ is the refined feature of the $i$th branch and $\alpha_i$ is the corresponding dynamic fusion weight. This mechanism adapts branch contributions to the global context and channel response of each input.

The fused feature is integrated by a $1 \times 1$ convolution and added to the residual mapping of the trunk feature. Unlike direct concatenation, this design allows the decoder to distinguish reusable structural detail from degradation responses inherited from shallow layers. The three receptive fields also provide complementary context for objects of different scales. SKCFM therefore uses skip details selectively while filtering redundant haze responses and strengthening multi-scale context representation.

Before selective fusion, both skip features are resized to match the decoder trunk and projected to a common channel dimension. BRCM then constructs complementary responses between the trunk and each skip feature. This step is useful because shallow features contain sharp boundaries but may also retain haze and noise. After gating, the trunk and the two refined skips enter $3 \times 3$, $5 \times 5$ and $7 \times 7$ convolutional branches. The different receptive fields provide context for structures with varied spatial extent.

SKFusion estimates branch-channel weights from globally pooled responses and applies Softmax normalisation across the three paths. The fused output is projected and added to a residual transform of the trunk. This structure offers two safeguards: BRCM screens the skip features before fusion, and SKFusion adjusts their relative importance afterwards. Their combination is intended to prevent the decoder from treating all skip information as equally reliable.

\begin{figure}[t]
\centering
\includegraphics[width=\linewidth]{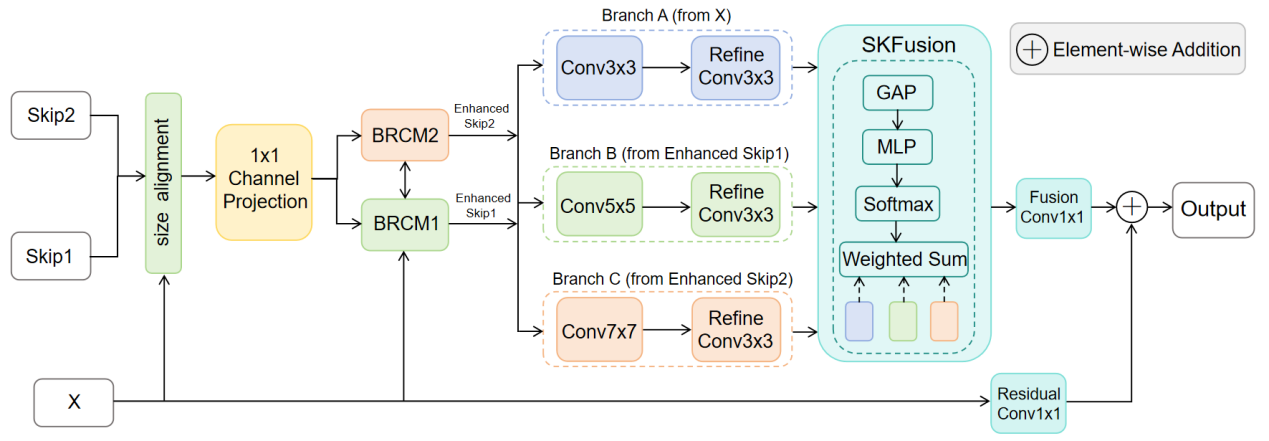}
\caption{Selective kernel complementary fusion module (SKCFM). BRCM first screens two skip features against the decoder trunk, after which SKFusion aggregates three receptive fields with content-dependent weights.}
\label{fig:5}
\end{figure}

\subsection{Loss functions}

DPSF-Net is optimised with three complementary losses: L1 loss, structure-aware channel-weighted perceptual (SCWP) loss and contrastive loss. L1 loss constrains pixel reconstruction, SCWP loss improves structural perceptual consistency using VGG19 features \cite{ref39}, \cite{ref40}, and contrastive loss follows physics-aware dehazing constraints \cite{ref41}. Because DCP is an auxiliary input, all losses are applied only to the restored RGB image.

\subsubsection{L1 loss}

L1 loss constrains the pixel-level difference between the restored image and the clear reference:

\begin{equation}
L_1 = \lVert J-G \rVert_1.
\end{equation}

where J is the restored RGB image and G is the clear reference image. This term preserves overall brightness, colour and structure.

\subsubsection{SCWP loss}

To improve edge, texture and structure recovery, we use SCWP loss. VGG19 extracts multi-layer perceptual features, and channel weights are generated from the spatial standard deviation of GT perceptual features:

\begin{equation}
w_{l,c}=\operatorname{Softmax}_c\left(\operatorname{Std}_{h,w}\left(\phi_l(G)_{c,h,w}\right)\right).
\end{equation}

where $\phi_l$ denotes the $l$th-layer VGG19 feature and $w_{l,c}$ is the weight of channel $c$ at layer $l$. Channels with higher spatial standard deviation contain richer structural and texture variation and are therefore assigned greater perceptual weight. SCWP loss is defined as:

\begin{equation}
L_{\mathrm{SCWP}} = \lambda_s \sum_l \sum_c w_{l,c}\frac{1}{H_lW_l}\left\lVert \phi_l(J)_c-\phi_l(G)_c\right\rVert_1.
\end{equation}

SCWP loss increases the contribution of perceptual channels with rich structural variation, improving recovery of building edges, road textures and object contours.

\subsubsection{Contrastive loss}

To strengthen the ability of PGFAM features to distinguish clear images from hazy inputs, we introduce contrastive loss. During training, PGFAM generates two-scale anchor features $A_4$ and $A_2$. Clear images are treated as positive samples $P$, and hazy RGB images are treated as negative samples $N$. For a single anchor $A$, the contrastive loss is:

\begin{equation}
L_{\mathrm{con}}(A)=\lVert A-P\rVert_2^2+\max\left(m-\lVert A-N\rVert_2^2,0\right).
\end{equation}

Here, $P$ denotes the projected feature of the clear image, $N$ denotes the projected feature of the hazy RGB image, and $m$ is the margin. The final contrastive loss sums the two scale-specific losses:

\begin{equation}
L_{\mathrm{contrast}} = \lambda_c\left[L_{\mathrm{con}}(A_4)+L_{\mathrm{con}}(A_2)\right].
\end{equation}

This term pulls anchor features towards clear-image features and pushes them away from hazy-image features. The total loss is:

\begin{equation}
L_{\mathrm{total}} = L_1 + L_{\mathrm{SCWP}} + L_{\mathrm{contrast}}.
\end{equation}

L1 loss maintains pixel-level reconstruction accuracy, SCWP loss strengthens structurally informative perceptual channels, and contrastive loss improves discrimination between clear and hazy features.

The three objectives operate at complementary levels. Pixel supervision constrains the overall reconstruction, SCWP emphasises perceptual channels with greater structural variation, and contrastive supervision regularises intermediate anchor features. The loss design is consistent with the network architecture: the output should remain faithful at the pixel level, preserve important edges and textures, and move intermediate representations away from haze-related patterns.

\section{Experiments}

\subsection{Experimental settings}

\subsubsection{Datasets}

We evaluate DPSF-Net on the real-world RRSHID dataset \cite{ref19} and four public remote sensing dehazing benchmarks: RSID \cite{ref21}, RICE1 \cite{ref42}, RICE2 \cite{ref42} and StateHaze1K-thick \cite{ref43}. These datasets cover synthetic haze, cloud contamination, thick cloud and real-world non-uniform haze, enabling evaluation across different degradation types.

RSID contains 1000 hazy/clear image pairs, with 900 for training and 100 for testing. RICE1 contains 500 cloudy/cloud-free pairs (400/100 train/test), and RICE2 contains 736 thick-cloud and shadow-degraded pairs (588/147 train/test). StateHaze1K-thick contains 400 image pairs, with 320 for training, 45 for testing and 35 for validation. RRSHID targets real remote sensing haze and better reflects complex atmosphere, non-uniform haze and colour distortion.

The benchmarks serve different evaluation purposes. RRSHID measures restoration under real atmospheric degradation and is the principal dataset for comparison and ablation. RSID evaluates transfer to paired synthetic haze. RICE1 and RICE2 introduce cloud-related degradation with different severity, while StateHaze1K-thick focuses on dense haze. Evaluating these datasets together helps distinguish performance on the target real-world setting from broader cross-dataset behaviour. The same metric definitions are used throughout to keep the comparisons consistent.

\subsubsection{Implementation details}

All experiments were conducted on a single NVIDIA RTX 5880 GPU. Input images were resized to $256 \times 256$, the batch size was 4 and training ran for 300 epochs. We used the Adam optimiser \cite{ref44} with momentum parameters $\beta_1=0.9$ and $\beta_2=0.999$. The initial learning rate was $5 \times 10^{-4}$, and CosineAnnealingLR \cite{ref45} decayed it to $1 \times 10^{-8}$.

The network uses a five-stage encoder-decoder design with embedding dimensions [24, 48, 96, 48, 24] and stage depths [8, 8, 16, 8, 8]. DPSF-Net differs from MCAF-Net by using RGB-DCP 6-channel inputs and by combining L1, SCWP and contrastive losses to improve recovery from non-uniform haze, colour shift and detail degradation.

For each hazy RGB image, the corresponding DCP map is prepared before the network forward pass and concatenated with the RGB channels. The restored RGB output is used for validation metrics, whereas the DCP channels remain auxiliary inputs. During training, contrastive supervision is applied to anchor features generated at the two decoder scales. This separation ensures that the reported image-quality metrics reflect the restored colour image rather than the auxiliary representation.

The experimental configuration is intentionally compact. It uses the same five-stage width schedule as the lightweight baseline and introduces additional computation only where it supports the proposed design goals. Fourier-domain interaction is confined to one quarter of the SFRIB channels. PGFAM uses depthwise and pointwise projections, and SKCFM performs selective fusion after channel alignment. This setup allows the ablation study to relate quality changes to identifiable architectural additions.

\subsubsection{Evaluation metrics}

We report four full-reference image-quality metrics: PSNR, SSIM \cite{ref46}, MSE and LPIPS \cite{ref47}. Higher PSNR and SSIM indicate better pixel reconstruction and structural preservation, whereas lower MSE and LPIPS indicate smaller reconstruction error and better perceptual quality. These metrics are complementary: PSNR and MSE emphasise pixel fidelity, SSIM measures structural similarity, and LPIPS evaluates perceptual discrepancy in a learned feature space. We therefore interpret improvements across multiple metrics rather than relying on a single score.

The four metrics can occasionally favour different models. For example, a method may obtain a lower pixel error while producing less natural local contrast, or it may improve perceptual similarity without minimising MSE. We therefore report the complete metric set and discuss such cases explicitly. This practice is particularly relevant for remote sensing scenes, where small boundaries and colour consistency can matter even when global reconstruction errors are similar.

\subsection{Experimental Results}

\subsubsection{Real-world RRSHID evaluation}

We first evaluate DPSF-Net on RRSHID, which includes thin, moderate and thick haze. Table 1 reports quantitative results against representative methods, and Fig. 6 shows visual comparisons.

In the thin-haze subset, DPSF-Net achieves 24.09 dB PSNR, 0.6825 SSIM, 0.0056 MSE and 0.1341 LPIPS. Relative to MCAF-Net, PSNR increases by 0.77 dB, SSIM by 0.0589 and LPIPS decreases from 0.4023 to 0.1341, indicating stronger structural and perceptual restoration.

For moderate haze, DPSF-Net reaches 24.17 dB PSNR and 0.6973 SSIM, exceeding MCAF-Net by 0.57 dB and 0.0390. MSE remains 0.0063, whereas LPIPS decreases from 0.3799 to 0.1654, showing improved perceptual consistency without sacrificing pixel-error stability.

For thick haze, DPSF-Net obtains 25.63 dB PSNR and 0.7508 SSIM, improving on MCAF-Net by 0.23 dB and 0.0287. Although MSE is slightly higher than that of MCAF-Net , LPIPS decreases from 0.3942 to 0.1536, suggesting better perceptual restoration in severe haze.

Averaged across RRSHID, DPSF-Net achieves 24.63 dB PSNR, 0.7102 SSIM, 0.0053 MSE and 0.1510 LPIPS. Compared with MCAF-Net, it improves PSNR by 0.52 dB and SSIM by 0.0422 while reducing LPIPS by more than half. The consistent improvement across thin, moderate and thick subsets suggests that the model is not specialised to a single haze density. The particularly large LPIPS reduction indicates that its gains extend beyond pixel-level fitting and remain visible in perceptually relevant structures.

The visual results in Fig. 6 are consistent with the quantitative trends. DCP often darkens images and introduces colour distortion, whereas FFA-Net, 4KDehazing and DehazeFormer may leave residual haze or over-enhanced regions. PhDNet and MCAF-Net recover many ground structures, but blurred details and colour shifts remain in complex areas. DPSF-Net produces clearer building edges, road textures and object contours, with less residual haze and more natural colour.

The RRSHID results also reveal differences between the evaluation metrics. On the thick-haze subset, MCAF-Net obtains a marginally lower MSE than DPSF-Net, whereas DPSF-Net produces higher PSNR and SSIM and a substantially lower LPIPS. This pattern suggests that minimising average pixel error alone does not fully describe the restoration quality of severe real-world haze. The lower LPIPS is consistent with the visual recovery of roads, rooftops and object boundaries. Across all three haze levels, the gains are strongest for SSIM and LPIPS, which supports the emphasis on structural and perceptual recovery.

The average RRSHID results provide a useful summary of robustness across haze density. DPSF-Net improves PSNR and SSIM while reducing LPIPS from 0.3921 to 0.1510 relative to MCAF-Net. The improvement is not confined to a single subset. This consistency indicates that the auxiliary prior and frequency branch remain useful as haze severity changes. It also suggests that selective skip fusion limits the reintroduction of residual degradation during reconstruction.

\begin{table*}[t]
\centering
\caption{Performance comparison on RRSHID under different haze levels.}
\label{tab:1}
\scriptsize
\resizebox{\textwidth}{!}{%
\begin{tabular}{lrrrrrrrrrrrrrrrr}
\toprule
\textbf{Method}  &  \multicolumn{4}{c}{\textbf{RRSHID-thin}}  &  \multicolumn{4}{c}{\textbf{RRSHID-moderate}}  &  \multicolumn{4}{c}{\textbf{RRSHID-thick}}  &  \multicolumn{4}{c}{\textbf{RRSHID-average}} \\
\cmidrule(lr){2-5}\cmidrule(lr){6-9}\cmidrule(lr){10-13}\cmidrule(lr){14-17}
  &  \textbf{PSNR}  &  \textbf{SSIM}  &  \textbf{MSE}  &  \textbf{LPIPS}  &  \textbf{PSNR}  &  \textbf{SSIM}  &  \textbf{MSE}  &  \textbf{LPIPS}  &  \textbf{PSNR}  &  \textbf{SSIM}  &  \textbf{MSE}  &  \textbf{LPIPS}  &  \textbf{PSNR}  &  \textbf{SSIM}  &  \textbf{MSE}  &  \textbf{LPIPS} \\
\midrule
\textbf{DCP\cite{ref7}} & 18.46 & 0.4564 & 0.0192 & 0.4851 & 17.80 & 0.4856 & 0.0238 & 0.4700 & 18.39 & 0.4843 & 0.0208 & 0.4996 & 18.22 & 0.4754 & 0.0213 & 0.4849 \\
\textbf{FFA-Net\cite{ref14}} & 17.08 & 0.4452 & 0.0326 & 0.5761 & 17.40 & 0.5385 & 0.0346 & 0.5450 & 16.71 & 0.4792 & 0.0377 & 0.5573 & 17.06 & 0.4876 & 0.0350 & 0.5595 \\
\textbf{GridDehazeNet\cite{ref15}} & 22.77 & 0.6145 & 0.0069 & 0.4123 & 22.62 & 0.6468 & 0.0083 & 0.3833 & 23.96 & 0.7112 & 0.0061 & 0.3947 & 23.12 & 0.6575 & 0.0071 & 0.3968 \\
\textbf{4KDehazing\cite{ref18}} & 22.83 & 0.6177 & 0.0063 & 0.4352 & 22.47 & 0.6505 & 0.0083 & 0.4590 & 22.55 & 0.6912 & 0.0099 & 0.4754 & 22.62 & 0.6531 & 0.0082 & 0.4565 \\
\textbf{SCAnet\cite{ref29}} & 18.37 & 0.4718 & 0.0200 & 0.4827 & 18.11 & 0.0538 & 0.0210 & 0.4962 & 19.07 & 0.5966 & 0.0180 & 0.4734 & 18.52 & 0.3741 & 0.0195 & 0.4841 \\
\textbf{Trinity-Net\cite{ref21}} & 20.51 & 0.5728 & 0.0120 & 0.4578 & 22.46 & 0.5728 & 0.0085 & 0.4314 & 24.11 & \underline{0.7234} & 0.0058 & 0.4206 & 22.36 & 0.6230 & 0.0060 & 0.4366 \\
\textbf{DehazeFormer\cite{ref16}} & 22.74 & 0.6005 & 0.0071 & 0.4609 & 23.06 & 0.6137 & 0.0076 & 0.4495 & 24.69 & 0.7143 & 0.0051 & 0.4377 & 23.50 & 0.6428 & 0.0066 & 0.4494 \\
\textbf{PCSformer\cite{ref26}} & 21.83 & 0.5427 & 0.0088 & 0.4963 & 22.09 & 0.5984 & 0.0092 & 0.5021 & 23.71 & 0.6547 & 0.0055 & 0.4996 & 22.54 & 0.5996 & 0.0070 & 0.4993 \\
\textbf{PhDnet\cite{ref30}} & 22.64 & 0.6054 & 0.0072 & 0.4658 & 22.92 & 0.6448 & \underline{0.0075} & 0.4019 & 24.28 & 0.6996 & 0.0053 & 0.3993 & 23.28 & 0.6499 & 0.0067 & 0.4223 \\
\textbf{MCAF-Net\cite{ref19}} & \underline{23.32} & \underline{0.6236} & \underline{0.0059} & \underline{0.4023} & \underline{23.60} & \underline{0.6583} & \textbf{0.0063} & \underline{0.3799} & \underline{25.40} & 0.7221 & \textbf{0.0040} & \underline{0.3942} & \underline{24.11} & \underline{0.6680} & \underline{0.0054} & \underline{0.3921} \\
\textbf{DPSF-Net} & \textbf{24.09} & \textbf{0.6825} & \textbf{0.0056} & \textbf{0.1341} & \textbf{24.17} & \textbf{0.6973} & \textbf{0.0063} & \textbf{0.1654} & \textbf{25.63} & \textbf{0.7508} & \underline{0.0041} & \textbf{0.1536} & \textbf{24.63} & \textbf{0.7102} & \textbf{0.0053} & \textbf{0.1510} \\
\bottomrule
\end{tabular}%
}
\end{table*}

\begin{figure*}[!t]
\centering
\includegraphics[width=\textwidth]{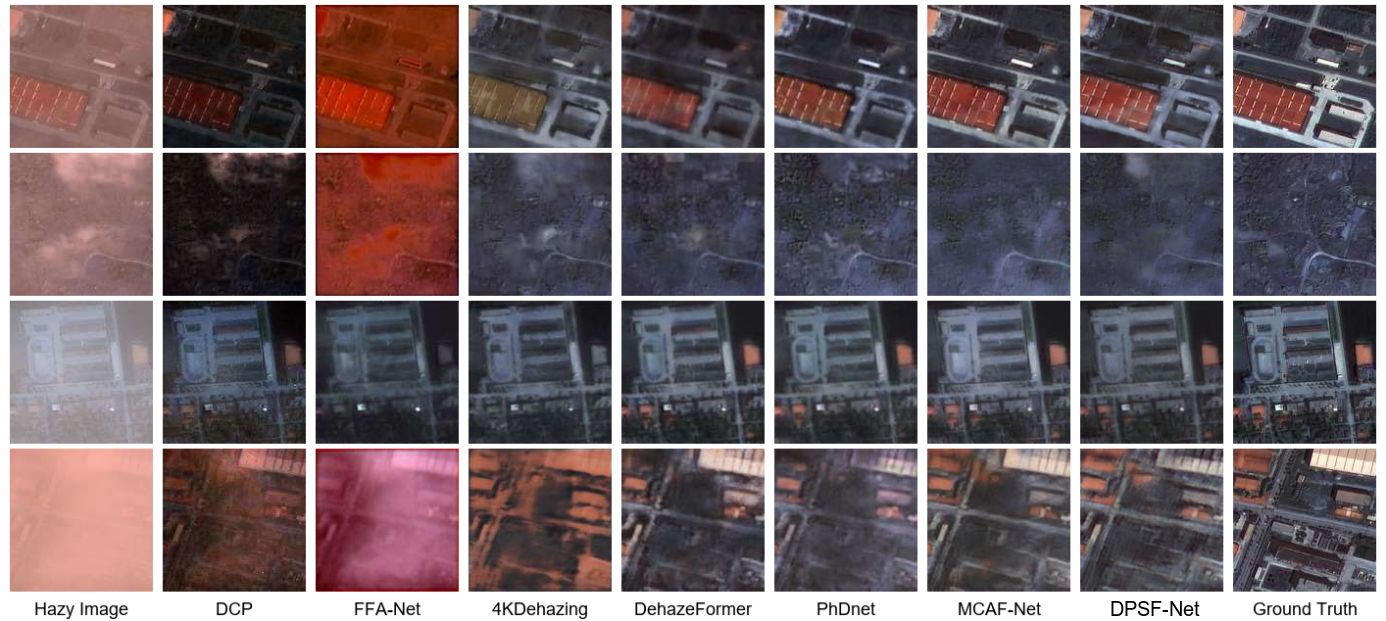}
\caption{Visual comparisons on the real-world RRSHID dataset. DPSF-Net removes residual haze while preserving road boundaries, rooftops and object contours across different haze levels.}
\label{fig:6}
\end{figure*}

\subsubsection{Synthetic RSID evaluation}

To test cross-dataset generalisation, we further evaluate DPSF-Net on RSID, RICE1, RICE2 and StateHaze1K-thick. These benchmarks differ from RRSHID in degradation source and imaging condition. Table 2 reports quantitative results, and Fig. 7 gives visual examples.

On RSID, DPSF-Net obtains 25.96 dB PSNR, 0.9561 SSIM, 0.0030 MSE and 0.0210 LPIPS. It slightly improves PSNR and SSIM over MCAF-Net and halves LPIPS, indicating stronger perceptual restoration.

On RICE1, DPSF-Net reaches 35.92 dB PSNR and 0.9850 SSIM, exceeding MCAF-Net while reducing MSE to 0.0004 and LPIPS to 0.0069. This suggests reliable recovery under relatively regular cloud cover.

On RICE2, DPSF-Net achieves 34.70 dB PSNR, 0.9069 SSIM and 0.0987 LPIPS. Although its MSE is slightly higher than that of MCAF-Net, the PSNR, SSIM and LPIPS gains indicate better structure and perceptual consistency under thick cloud and shadow degradation.

On StateHaze1K-thick, DPSF-Net improves PSNR from 22.16 dB to 22.72 dB relative to MCAF-Net and reduces LPIPS from 0.1453 to 0.0527, while maintaining the same SSIM of 0.9308.

Fig. 7 shows that DPSF-Net generally yields clearer edges, more natural colour and fewer residual haze artefacts than comparison methods across the four synthetic benchmarks. The cross-dataset results are important because the degradation patterns differ substantially from RRSHID. Improvements on these benchmarks indicate that RGB-DCP guidance and spatial-frequency modelling remain useful when the haze source, cloud thickness and surface composition vary.

The synthetic benchmarks expose several complementary behaviours. On RSID, DPSF-Net improves all four metrics over MCAF-Net, including a reduction in LPIPS from 0.0421 to 0.0210. On RICE1, the method obtains the strongest SSIM, MSE and LPIPS among the reported values, although DehazeFormer has a slightly higher PSNR. On RICE2, DPSF-Net improves PSNR, SSIM and LPIPS, while MCAF-Net retains the lowest MSE. On StateHaze1K-thick, DPSF-Net increases PSNR and reduces both MSE and LPIPS, with SSIM tied with MCAF-Net. These results reinforce the need to consider several metrics together.

Qualitative comparisons further show that cloud-like degradation requires more than global brightness correction. Restored images should recover land-cover boundaries without introducing artificial colour transitions. The combination of local interaction, global frequency context and prior-guided calibration is intended to address this balance. The cross-dataset results do not imply universal generalisation, but they show that the proposed modules remain useful outside the principal RRSHID setting.

\begin{table*}[!t]
\centering
\caption{Performance comparison on multiple haze benchmarks.}
\label{tab:2}
\scriptsize
\resizebox{\textwidth}{!}{%
\begin{tabular}{lrrrrrrrrrrrrrrrr}
\toprule
\textbf{Method}  &  \multicolumn{4}{c}{\textbf{RSID}}  &  \multicolumn{4}{c}{\textbf{RICE1}}  &  \multicolumn{4}{c}{\textbf{RICE2}}  &  \multicolumn{4}{c}{\textbf{Statehaze1K-thick}} \\
\cmidrule(lr){2-5}\cmidrule(lr){6-9}\cmidrule(lr){10-13}\cmidrule(lr){14-17}
  &  \textbf{PSNR}  &  \textbf{SSIM}  &  \textbf{MSE}  &  \textbf{LPIPS}  &  \textbf{PSNR}  &  \textbf{SSIM}  &  \textbf{MSE}  &  \textbf{LPIPS}  &  \textbf{PSNR}  &  \textbf{SSIM}  &  \textbf{MSE}  &  \textbf{LPIPS}  &  \textbf{PSNR}  &  \textbf{SSIM}  &  \textbf{MSE}  &  \textbf{LPIPS} \\
\midrule
\textbf{DCP\cite{ref7}} & 13.87 & 0.6892 & 0.0538 & 0.2615 & 15.68 & 0.6860 & 0.0491 & 0.3766 & 14.10 & 0.3898 & 0.0656 & 0.6234 & 17.87 & 0.8481 & 0.0169 & 0.1879 \\
\textbf{FFA-Net\cite{ref14}} & 18.31 & 0.8582 & 0.0204 & 0.1449 & 23.73 & 0.9068 & 0.0096 & 0.1885 & 17.77 & 0.6261 & 0.0323 & 0.5378 & 19.45 & 0.9023 & 0.0117 & 0.2357 \\
\textbf{GridDehazeNet\cite{ref15}} & 23.50 & 0.9383 & 0.0080 & 0.0541 & 33.45 & 0.9766 & 0.0010 & 0.0405 & 31.54 & 0.8839 & 0.0014 & \underline{0.1942} & 20.51 & 0.9097 & 0.0089 & \underline{0.1361} \\
\textbf{4KDehazing\cite{ref18}} & 23.61 & 0.9415 & 0.0053 & 0.0693 & 27.54 & 0.9425 & 0.0024 & 0.0967 & 25.21 & 0.8604 & 0.0087 & 0.3704 & 20.75 & 0.7696 & 0.0085 & 0.1674 \\
\textbf{Trinity-Net\cite{ref21}} & 23.60 & 0.9322 & 0.0060 & 0.0692 & 23.46 & 0.8796 & 0.0100 & 0.0512 & 18.81 & 0.7781 & 0.0200 & 0.2287 & 20.43 & 0.8056 & 0.0086 & 0.1764 \\
\textbf{PSMB-Net\cite{ref24}} & 25.64 & 0.9447 & \textbf{0.0030} & 0.0537 & 31.32 & 0.9444 & 0.0006 & 0.0428 & 30.21 & 0.8756 & \underline{0.0010} & 0.1983 & 21.55 & 0.8489 & 0.0084 & 0.1642 \\
\textbf{DehazeFormer\cite{ref16}} & 25.04 & 0.9393 & 0.0040 & 0.0557 & \textbf{36.15} & 0.9794 & \underline{0.0005} & 0.0460 & 34.54 & 0.8875 & \underline{0.0010} & 0.2035 & 22.02 & \underline{0.9306} & 0.0063 & 0.1707 \\
\textbf{PCSformer\cite{ref26}} & 23.03 & 0.9159 & 0.0063 & 0.0718 & 35.17 & 0.9564 & 0.0006 & 0.0421 & 33.80 & 0.8781 & 0.0011 & 0.2418 & 20.20 & 0.8125 & 0.0097 & 0.1734 \\
\textbf{PhDnet\cite{ref30}} & 25.15 & 0.9406 & 0.0040 & 0.0612 & 35.50 & 0.9578 & \underline{0.0005} & 0.0464 & 33.44 & 0.8877 & 0.0011 & 0.2277 & 21.99 & 0.8294 & 0.0064 & 0.1721 \\
\textbf{MCAF-Net\cite{ref19}} & \underline{25.89} & \underline{0.9531} & \underline{0.0032} & \underline{0.0421} & 35.79 & \underline{0.9814} & \underline{0.0005} & \underline{0.0383} & \underline{34.59} & \underline{0.8948} & \textbf{0.0009} & 0.1964 & \underline{22.16} & \textbf{0.9308} & \underline{0.0062} & 0.1453 \\
\textbf{DPSF-Net} & \textbf{25.96} & \textbf{0.9561} & \textbf{0.0030} & \textbf{0.0210} & \underline{35.92} & \textbf{0.9850} & \textbf{0.0004} & \textbf{0.0069} & \textbf{34.70} & \textbf{0.9069} & 0.0012 & \textbf{0.0987} & \textbf{22.72} & \textbf{0.9308} & \textbf{0.0054} & \textbf{0.0527} \\
\bottomrule
\end{tabular}%
}
\vspace{0.4em}
\centering
\includegraphics[width=0.94\textwidth]{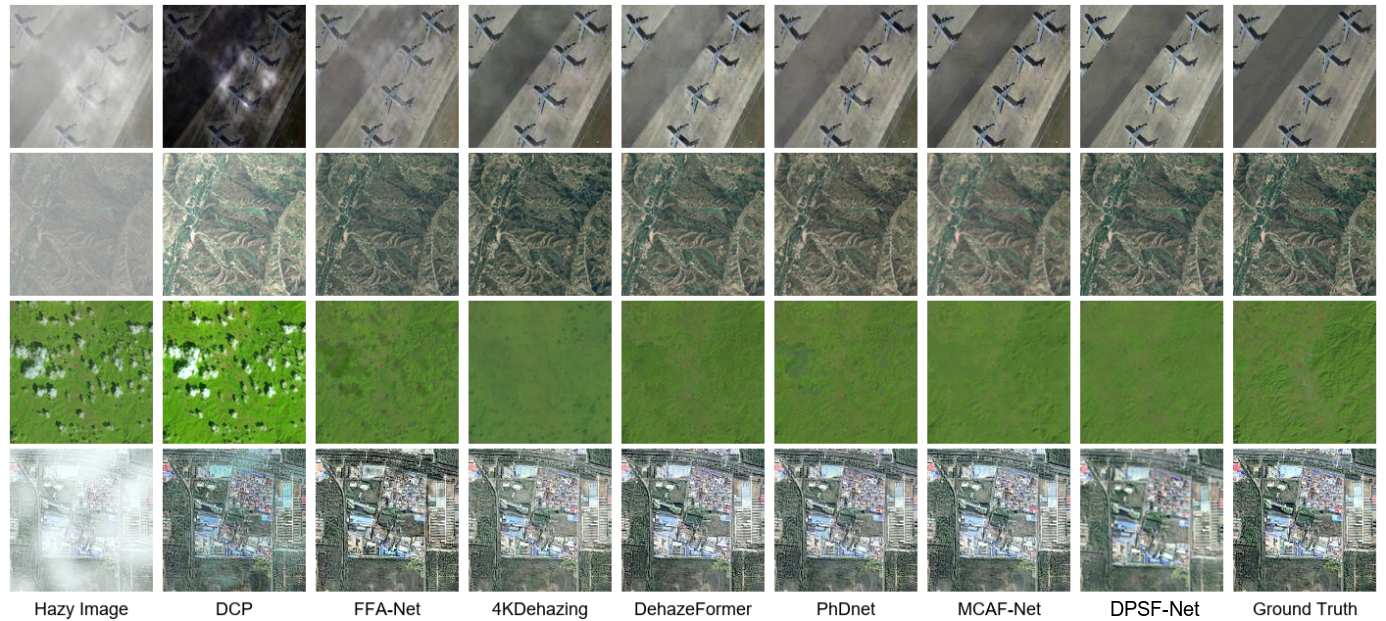}
\captionof{figure}{Visual comparisons on RSID, RICE1, RICE2 and StateHaze1K-thick. DPSF-Net produces clearer structures and more consistent colours across synthetic haze and cloud-degradation conditions.}
\label{fig:7}
\end{table*}

\subsection{Ablation studies}

We conduct stepwise ablation on RRSHID-thick to assess the effect of each component. MCAF-Net is used as the baseline, and FourierUnit, BRCM, SKFusion and 6-channel DCP input are added sequentially. Quantitative results are shown in Table 3, and visual examples are shown in Fig. 8.

Adding FourierUnit increases PSNR from 24.66 dB to 25.16 dB and reduces MSE from 0.0064 to 0.0058, indicating improved modelling of global haze degradation. Although SSIM decreases slightly, the lower reconstruction error supports the value of frequency-domain context.

Adding BRCM further increases PSNR to 25.26 dB and reduces MSE to 0.0049, suggesting that bidirectional residual complementary gating helps suppress redundant haze in skip features. With SKFusion, PSNR and SSIM rise to 25.50 dB and 0.7495, showing that dynamic scale selection improves feature fusion.

After introducing the 6-channel DCP input, DPSF-Net achieves the best overall result: 25.63 dB PSNR, 0.7508 SSIM and 0.0041 MSE. Compared with the baseline, PSNR increases by 0.97 dB and MSE decreases by 0.0023. The final model has 669.8K parameters and 18.27G FLOPs, remaining computationally compact.

The visual ablation in Fig. 8 shows progressively reduced residual haze and clearer surface textures as the proposed components are added. The full model gives the closest colour, structure and haze-removal effect to the reference image.

The component ablation suggests a staged interpretation of the architecture. FourierUnit first improves the modelling of spatially extended degradation and increases PSNR by 0.50 dB over the baseline. BRCM then lowers MSE from 0.0058 to 0.0049, indicating that complementary gating helps regulate skip-feature reuse. SKFusion improves SSIM from 0.7374 to 0.7495, which is consistent with more selective multi-scale aggregation. Finally, the 6-channel DCP input yields the strongest overall quality. The sequence shows that the prior, frequency and fusion mechanisms make complementary contributions.

The computational changes are also informative. Replacing the pointwise branch with FourierUnit does not produce a large model, and the final configuration remains below the FLOPs of the baseline. BRCM and SKFusion introduce additional parameters, but the final network still contains only 669.8K parameters. The ablation therefore supports the intended trade-off: the revised modules improve restoration quality without converting the lightweight baseline into a high-capacity model.

\begin{table}[t]
\centering
\caption{Ablation results for core modules.}
\label{tab:3}
\small
\resizebox{\linewidth}{!}{%
\begin{tabular}{lrrrrr}
\toprule
\textbf{Model}  &  \textbf{PSNR}  &  \textbf{SSIM}  &  \textbf{MSE}  &  \textbf{Params}  &  \textbf{FLOPs} \\
\midrule
\textbf{Baseline} & 24.66 & 0.7393 & 0.0064 & \textbf{558.1K} & 19.82G \\
\textbf{+FourierUnit} & 25.16 & 0.7366 & 0.0058 & \underline{619.5K} & \textbf{14.74G} \\
\textbf{+BRCM} & 25.26 & 0.7374 & \underline{0.0049} & 629.8K & \underline{15.34G} \\
\textbf{+SKFusion} & \underline{25.50} & \underline{0.7495} & 0.0050 & 665.6K & 17.32G \\
\textbf{+6ch DCP} & \textbf{25.63} & \textbf{0.7508} & \textbf{0.0041} & 669.8K & 18.27G \\
\bottomrule
\end{tabular}%
}
\end{table}

\begin{figure}[t]
\centering
\includegraphics[width=\linewidth]{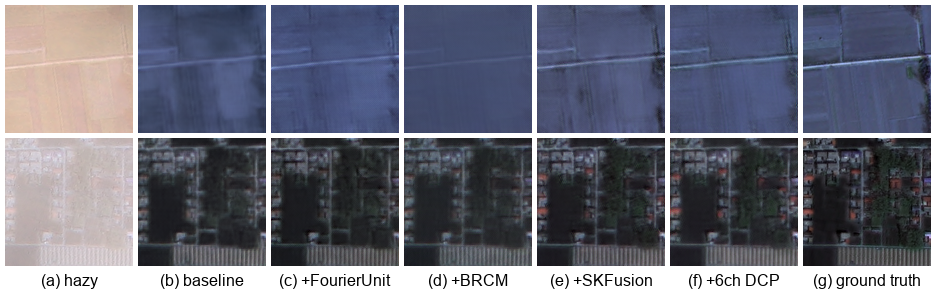}
\caption{Visual comparison of component ablations. FourierUnit, BRCM, SKFusion and DCP guidance progressively reduce residual haze and improve structural recovery.}
\label{fig:8}
\end{figure}

We also ablate the loss functions on RRSHID-thick. Table 4 and Fig. 9 compare $L_1$, $L_1 + L_{\mathrm{SCWP}}$, $L_1 + L_{\mathrm{contrast}}$, and the full $L_1 + L_{\mathrm{SCWP}} + L_{\mathrm{contrast}}$ objective.

With only $L_1$ loss, the model reaches 24.06 dB PSNR, 0.7182 SSIM and 0.0059 MSE, showing that pixel supervision alone is insufficient for fine structural and perceptual recovery. Adding $L_{\mathrm{SCWP}}$ increases PSNR to 24.90 dB and SSIM to 0.7455 while reducing MSE to 0.0053, indicating stronger edge, texture and object-structure restoration.

Using $L_1 + L_{\mathrm{contrast}}$ increases PSNR and SSIM to 24.47 dB and 0.7296, with MSE reduced to 0.0053. Contrastive loss improves the discrimination of clear structures, although its effect on structural preservation is weaker than SCWP when each is paired with $L_1$ alone.

The full $L_1 + L_{\mathrm{SCWP}} + L_{\mathrm{contrast}}$ objective performs best, reaching 25.63 dB PSNR, 0.7508 SSIM and 0.0041 MSE. Compared with $L_1$ alone, PSNR increases by 1.57 dB, SSIM by 0.0326 and MSE decreases by 0.0018. Fig. 9 shows that the full loss removes residual haze more effectively, restores sharper roads and object contours, and produces colours closer to the reference.

The loss ablation clarifies the role of the auxiliary objectives. SCWP provides the larger individual improvement when added to $L_1$, which is consistent with its emphasis on structurally informative perceptual channels. Contrastive supervision produces a smaller but measurable gain by separating anchor representations from haze-related features. When the two terms are combined, the network obtains the strongest result across all three metrics. The objectives therefore act cooperatively rather than redundantly.

\begin{table}[t]
\centering
\caption{Ablation results for loss functions.}
\label{tab:4}
\small
\resizebox{\linewidth}{!}{%
\begin{tabular}{lrrr}
\toprule
\textbf{Loss function}  &  \textbf{PSNR}  &  \textbf{SSIM}  &  \textbf{MSE} \\
\midrule
\textbf{$L_1$} & 24.06 & 0.7182 & 0.0059 \\
\textbf{$L_1 + L_{\mathrm{SCWP}}$} & \underline{24.90} & \underline{0.7455} & \underline{0.0053} \\
\textbf{$L_1 + L_{\mathrm{contrast}}$} & 24.47 & 0.7296 & \underline{0.0053} \\
\textbf{$L_1 + L_{\mathrm{SCWP}} + L_{\mathrm{contrast}}$} & \textbf{25.63} & \textbf{0.7508} & \textbf{0.0041} \\
\bottomrule
\end{tabular}%
}
\end{table}

\begin{figure}[t]
\centering
\includegraphics[width=\linewidth]{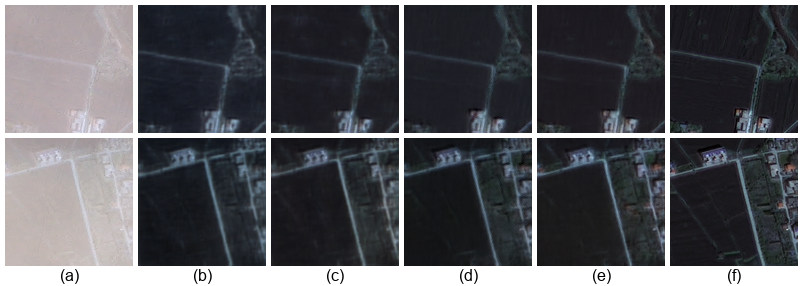}
\caption{Visual comparison of loss configurations: (a) hazy image, (b) $L_1$, (c) $L_1 + L_{\mathrm{contrast}}$, (d) $L_1 + L_{\mathrm{SCWP}}$, (e) $L_1 + L_{\mathrm{SCWP}} + L_{\mathrm{contrast}}$, and (f) ground truth. Combining perceptual and contrastive constraints yields the clearest boundaries and the most consistent restoration.}
\label{fig:9}
\end{figure}

Table 5 compares restoration quality and computational efficiency on RRSHID-thick. DPSF-Net achieves the best PSNR and SSIM among all methods, reaching 25.63 dB and 0.7508. Relative to MCAF-Net, it improves PSNR by 0.23 dB and SSIM by 0.0287.

DPSF-Net uses 669.8K parameters, far fewer than DehazeFormer, PCSformer, PhDNet and Trinity-Net. Its FLOPs are 18.27G, lower than MCAF-Net's 19.82G, and its inference time is comparable. The model therefore improves restoration quality while preserving low computational cost. This balance is relevant for remote sensing workflows, where large images and repeated scene processing can amplify even modest computational overhead. The results also show that global context modelling does not require a substantially larger backbone when frequency-domain interaction is introduced selectively.

\begin{table}[t]
\centering
\caption{Performance and complexity comparison with state-of-the-art methods.}
\label{tab:5}
\small
\resizebox{\linewidth}{!}{%
\begin{tabular}{lrrrrrr}
\toprule
\textbf{Method}  &  \textbf{PSNR}  &  \textbf{SSIM}  &  \textbf{MSE}  &  \textbf{Time (s)}  &  \textbf{Params}  &  \textbf{FLOPs} \\
\midrule
\textbf{FFA-Net} & 16.71 & 0.4792 & 0.0377 & 0.1200 & 4.456M & 287.8G \\
\textbf{GridDehazeNet} & 20.24 & 0.6312 & 0.0096 & 0.0699 & 955.7K & 85.72G \\
\textbf{4KDehazing} & 22.55 & 0.6912 & 0.0099 & 0.0549 & 34.55M & 105.8G \\
\textbf{SCANet} & 19.07 & 0.5966 & 0.0180 & 0.0700 & 2.39M & 258.6G \\
\textbf{Trinity-Net} & 24.11 & 0.7103 & 0.0058 & \textbf{0.0400} & 20.14M & 30.78G \\
\textbf{DehazeFormer} & 24.69 & 0.7143 & 0.0051 & 0.0831 & 1.205M & 39.76G \\
\textbf{PCSformer} & 23.71 & 0.6547 & 0.0055 & 0.0600 & 3.73M & 27.66G \\
\textbf{PhDNet} & 24.28 & 0.6996 & 0.0053 & 0.0642 & 10.03M & 33.24G \\
\textbf{MCAF-Net} & \underline{25.40} & \underline{0.7221} & \textbf{0.0040} & \underline{0.0498} & \textbf{558.1K} & \underline{19.82G} \\
\textbf{DPSF-Net} & \textbf{25.63} & \textbf{0.7508} & \underline{0.0041} & 0.0501 & \underline{669.8K} & \textbf{18.27G} \\
\bottomrule
\end{tabular}%
}
\end{table}

\subsection{Discussion}

The experiments indicate that the performance gain of DPSF-Net arises from complementary design choices rather than a single dominant component. FourierUnit improves global background modelling, whereas BRCM and SKFusion regulate the reuse of shallow details. DCP guidance adds a physically interpretable cue at the input stage, and PGFAM learns how strongly prior-related features should influence the current representation. The ablation results support this interpretation: each component produces an incremental improvement, and their combination yields the strongest restoration quality.

The comparison between RRSHID and the synthetic benchmarks also highlights the intended scope of the method. DPSF-Net is designed for remote sensing scenes with non-uniform haze, colour deviation and heterogeneous land cover. Its improvement on RRSHID suggests that explicit prior guidance remains useful under real atmospheric degradation. Its competitive performance on RSID, RICE1, RICE2 and StateHaze1K-thick further indicates that the spatial-frequency representation transfers across several degradation patterns. Nevertheless, these results should not be interpreted as evidence of universal robustness.

The results also illustrate why the method uses DCP as guidance rather than as a fixed restoration rule. Direct DCP processing can produce dark images and colour distortion, as visible in the qualitative comparisons. DPSF-Net instead embeds the prior and combines it with learned representations. PGFAM can modulate the influence of prior-related features, while the residual reconstruction pathway preserves information from the RGB input. This design retains the interpretability of a physical cue without forcing every scene region to satisfy the same assumption.

From an efficiency perspective, DPSF-Net occupies a useful middle ground. It improves perceptual and structural quality without relying on a large transformer backbone or a computationally heavy multi-stage pipeline. The modest parameter count is relevant for repeated processing of high-resolution remote sensing scenes. However, the current experiments evaluate cropped inputs and full-reference metrics. Additional deployment studies are required before drawing conclusions about operational throughput on very large images.

\subsubsection{Limitations}

Several limitations remain. First, DCP is only an approximate physical cue and may be unreliable over bright surfaces, dense clouds or scenes with strong cloud-haze overlap. PGFAM can attenuate such responses, but it cannot fully recover information that is absent from the input. Second, the current evaluation focuses on full-reference image restoration. The practical value of dehazing should also be assessed through downstream remote sensing tasks, including object detection, change detection and semantic segmentation. Third, the present model processes RGB imagery and does not yet exploit temporal observations or additional spectral bands.

\subsubsection{Future work}

Future work will investigate scene-adaptive prior selection and semantic context modelling to reduce sensitivity to difficult bright regions. Multi-temporal, multispectral and multi-sensor inputs may provide complementary evidence when fine structures are severely obscured. It will also be important to evaluate whether perceptual improvements translate into measurable gains for downstream interpretation tasks and to examine deployment on large remote sensing images under practical memory constraints.

\section{Conclusion}

This study presents DPSF-Net, a dual-prior spatial-frequency network for real-world remote sensing image dehazing. The model combines RGB-DCP embedding, FourierUnit-based spatial-frequency interaction, prior-guided attention and selective skip-feature fusion. Together, these components improve the recovery of local structures while accounting for spatially extended haze and colour deviation.

Experiments on RRSHID and four public benchmarks show that DPSF-Net achieves a favourable balance between restoration quality and computational cost. The ablation studies further clarify the contributions of FourierUnit, BRCM, SKFusion, DCP guidance, SCWP loss and contrastive supervision. These results support dual-prior spatial-frequency modelling as an effective direction for compact real-world RSID systems.

\bibliographystyle{IEEEtran}
\bibliography{citations}

\end{document}